\documentclass[lettersize,journal]{IEEEtran}
\usepackage{amsmath,amsfonts}
\usepackage{algorithmic}
\usepackage{algorithm}
\usepackage{array}
\usepackage[caption=false,font=normalsize,labelfont=sf,textfont=sf]{subfig}
\usepackage{textcomp}
\usepackage{stfloats}
\usepackage{url}
\usepackage{verbatim}
\usepackage{graphicx}
\usepackage{booktabs}
\usepackage[colorinlistoftodos,prependcaption,textsize=tiny]{todonotes}
\usepackage{lipsum}
\usepackage{bbm}
\usepackage{arydshln}
\usepackage{amsmath}
\usepackage{comment}
\usepackage{wrapfig}
\usepackage[numbers]{natbib}
\usepackage{xcolor}
\usepackage[colorlinks=true, urlcolor=blue, linkcolor=red]{hyperref}
\usepackage{soul}
\usepackage{hyperref}
\usepackage{enumitem,amssymb}
\newlist{todolist}{itemize}{2}
\setlist[todolist]{label=$\square$}

\usepackage{amsmath,amssymb,amsfonts}
\usepackage{graphicx}
\usepackage{textcomp}
\usepackage{xcolor}

\begin{document}

\title{Small but Fair! Fairness for Multimodal Human-Human and Robot-Human Mental Wellbeing Coaching}


\author{
\IEEEauthorblockN{Jiaee Cheong*}\thanks{$^{*}$equal contribution, alphabetical order}
\and
\IEEEauthorblockN{Micol Spitale*}
\and
\IEEEauthorblockN{Hatice Gunes}
}

\markboth{Journal of \LaTeX\ Class Files,~Vol.~14, No.~8, August~2021}%
{Shell \MakeLowercase{\textit{et al.}}: A Sample Article Using IEEEtran.cls for IEEE Journals}

\IEEEpubid{0000--0000/00\$00.00~\copyright~2021 IEEE}


\maketitle

\begin{abstract}
In recent years, the affective computing (AC) and human-robot interaction (HRI) research communities have put fairness at the centre of their research agenda. 
However, 
none of the existing work has addressed the problem of machine learning (ML) bias in HRI settings.
In addition, many of the current datasets for AC and HRI are `small', making ML bias and debias analysis challenging. 
This paper presents the first work to explore ML bias analysis and mitigation of three small multimodal datasets collected within both a human-human and robot-human wellbeing coaching settings. 
The contributions of this work includes:
i) being the first to explore the problem of ML bias and fairness within HRI settings; and
ii) providing a multimodal analysis evaluated via modelling performance and fairness metrics across both high and low-level features and proposing a simple and effective data augmentation strategy (MixFeat) to debias the small datasets presented within this paper; and 
iii) conducting extensive experimentation and analyses to reveal ML fairness insights unique to AC and HRI research in order to distill a set of recommendations to 
aid AC and HRI researchers to be more engaged with fairness-aware 
ML-based research.
%
%
\end{abstract}




\section{Introduction}

In recent years, the advancement in machine learning (ML), the availability of large-scale datasets and the enhancement in computing have led to the widespread use of machine-learning prediction systems in our society \cite{sarker2021machine}.  
However, the problem of bias in machine-learning based tools and systems are becoming an increasing source of concern \cite{barocas2017fairness}.
Such risks are also present in the field of affective computing as affect recognition tools are increasingly deployed in a wide range of high-stake use-cases 
such as mental wellbeing prediction \cite{su2020deep}
 and robotic mental wellbeing coaching \cite{churamani2022continual,spitale2023vita}.
The problem of bias is also becoming an increasingly greater source of concern within the human-robot interaction (HRI) research community \cite{claure2022fairness}.
Some of the fairness related concerns highlighted include fairness within a HRI teamwork context \cite{chang2021unfair}, robot navigation \cite{brandao2020fair} as well as HRI ethics and robot design \cite{ostrowski2022ethics}.  
However, this relatively nascent field has yet to consider the fairness-related challenges that occur due to the bias present in ML algorithms deployed within a HRI setting.

\begin{figure}
    \centering
    \includegraphics[width=0.7\columnwidth]{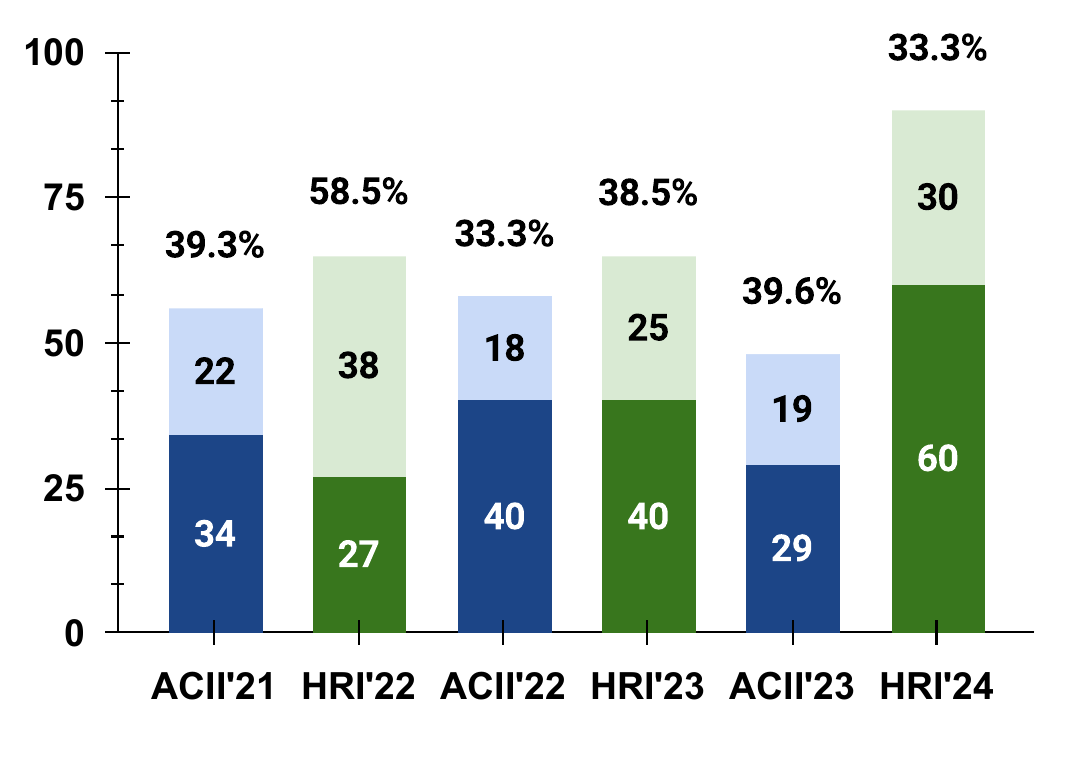}
    \vspace{-0.3cm}
    \caption{Proportion of small dataset papers accepted at ACII'21-'23 and HRI'22 - '24. Blue represents count for ACII. Green represents count for HRI.
    }
    \label{fig:small_dataset_ACII23}
\end{figure}

A wide range of fairness measures and bias mitigation techniques have been proposed to quantify and mitigate the bias present in machine learning models \cite{hort2022bias, cheong2021hitchhiker}.
As existing approaches chiefly focus on large datasets, they may not be effective for small datasets. 
However, most of the datasets currently available for affective computing (AC) application scenarios and within human-robot interaction (HRI) contexts are small, i.e., containing just a few hundred instances of data \cite{mathur2021modeling,chang2021unfair}.
%
Figure \ref{fig:small_dataset_ACII23} 
considers papers that have been published within the last three editions of the IEEE International Conference on Affective Computing \& Intelligent Interaction (ACII)  
and the ACM/IEEE international conference on Human-Robot Interaction (HRI)  
respectively. 
Both figures 
illustrate that papers focusing on small datasets typically represent 40\% to 60\% of the total papers accepted for presentation at the main conference track. 
Based on this, we consider any dataset that has less than 40 (median) subjects or 500 (median) samples `small'. We excluded papers that used large benchmark datasets such as AffectNet. 
Given the above, bias mitigation strategies developed within the mainstream ML fairness literature for large datasets may not work well on AC and HRI datasets.
Given that sensitive use cases such as wellbeing coaching \cite{spitale2022affective} and children's wellbeing assessment \cite{abbasi2022computational} are increasingly explored within both the AC and HRI research community,
there is an urgent need to develop bias mitigation methodologies that will work for small AC and HRI datasets.

Within our ACII 2023 paper \cite{cheong_acii}, we highlighted the ACII community’s attempt to be more ethically oriented as exemplified by the mandatory ethics impact statement to guard against the potential risks and harms that could be perpetuated by affect-related technology \footnote{https://acii-conf.net/2022/authors/submission-guidelines/}.
We hypothesised and showed that bias exists even for small datasets. Our experiments demonstrated that high-level features were often more informative and more reliable than low-level features and that a multimodal approach is often better than a unimodal approach across both performance and fairness metrics. We also proposed a simple, yet effective method that is able to mitigate against the bias present. 
Our intention is to contend that every analysis on small datasets should have a bias analysis section. 
In this work, within wellbeing context, we widen the impact of our previous work by extending the experiments towards small datasets in human-robot interaction scenarios.

We do so by introducing the first comprehensive work which explores the problem of bias in a small dataset within a high-stake and sensitive use of wellbeing coaching both within a dyadic human-human and robot-human mental wellbeing coaching small dataset setup. 
We investigate different data augmentation approaches to debias three small temporal multimodal mental wellbeing datasets. 
We further investigate the contribution of each individual modality (i.e., face, audio, verbal) and the importance of high and low-level features for data-driven applications. 
The main contributions of this paper are as follows. 
First, we conduct the first ML bias and fairness analysis in a HRI context by extending our human-human interaction (HHI) work in ACII'23 \cite{cheong_acii} to the HRI datasets in this paper.
Second, till now, the two HRI datasets have never been explored for ML-based wellbeing prediction.
Third, we provide a thorough multimodal analysis and a feature importance analysis evaluated using both performance and fairness metrics across three different wellbeing coaching datasets. 
Fourth, we experiment with different data augmentation strategies 
to reduce the bias in small dataset experimental settings. We compare our proposed method against a baseline data augmentation approach across both single and multiple modalities.
And last, we distill insights regarding bias and fairness within a small dataset setup and provide guidelines to assist affective computing and human-robot interaction researchers in future small dataset ML studies.
%
%

%
%
%

\section{Literature Review}


\subsection{Robotic Mental wellbeing Coaching}

Only a small number of studies have explored the use of robotic coaches to support mental wellbeing \cite{jeong2023deploying,  spitale2023vita, axelsson2023robotic}. 
For example, Jeong et al. \cite{jeong2023deploying} conducted a longitudinal study where Jibo robots provided positive psychology interventions to students in home settings over 7 days. The participants reported experiencing improved wellbeing, mood, and readiness to change, as well as developing an affinity for the robot.
Shi et al. \cite{shi2023evaluating} investigated how physical embodiment and personalization affect the perceived quality of text-to-speech (TTS) voices used for mindfulness exercises. They found that user-personalized TTS voices performed nearly as well as human voices, suggesting personalization can enhance the user's perception of TTS voice quality.
Spitale et al. \cite{spitale2023robotic} conducted a longitudinal study in which employees of a tech company interact with two different forms of robotic coaches that delivered positive psychology exercises over 4 weeks. Their results indicated that the specific form of the robotic coach may impact the users' perceptions and experiences.
Jeong et al. \cite{jeong2023robotic} explored how the robot's role (assistant, coach, or companion) affected the therapeutic alliance during wellbeing practice. They found that the companion robot was the most effective in building a positive relationship with users.
Axelsson et al. \cite{axelsson2023robotic} deployed a robotic mindfulness coach at a public cafe, where participants could join robot-led meditation sessions in a group setting. Their results suggest that participants perceived the robot-led mindfulness practice to be more helpful over time.
Spitale et al \cite{spitale2023vita} proposed a novel LLM-based multimodal system, namely VITA, that allows robotic coaches to autonomously adapt to the coachee's multi-modal behaviours (facial valence and speech duration) and deliver coaching exercises. 
Overall, this emerging body of research suggests that robotic coaches have potential to support mental wellbeing, but none of these previous studies have explored the bias and fairness of computational models embedded in robotic wellbeing coaches.


\subsection{Fairness in Mental wellbeing} 
Though recent attempts at applying ML for the investigation and understanding of mental health has been promising \cite{zhang2020multimodal},
there is only a handful of studies which have looked into bias in mental wellbeing prediction \cite{ryanfairness, bailey2021gender, zanna2022bias,cheong_gender_fairness}.
%
%
Zanna et al. \cite{zanna2022bias} conducted their experiments on data collected in the wild with a specific focus on anxiety prediction. 
Ryan et al. \cite{ryanfairness} proposed three categories of fairness definitions they deem relevant to mental health. 
Park et al. \cite{park2022fairness} analysed bias across gender in mobile mental health assessment and proposed an algorithmic impact remover to mitigate unwanted bias.
Bailey and Plumbley \cite{bailey2021gender} attempted to mitigate the gender bias present in the DAIC-WOZ dataset using data re-distribution.
\cite{cheong_gender_fairness} examined whether
bias exists in existing mental health datasets and
algorithms and provided practical suggestions to avoid hampering bias mitigation efforts in ML for mental health.
However, all of the existing works consist of relatively large datasets (more than 500 samples or more than 40 subjects) which differ from our small dataset setup.
In addition, no investigation has specifically looked into the problem of bias 
in the context of a human-human 
and robot-human mental wellbeing coaching.
%

\subsection{ML Bias and Fairness in HRI}
There is minimal work examining ML bias in the context of HRI. Most of the existing work on bias and fairness in HRI relate to fairness-related considerations within a HRI setting \cite{londono2022fairness,ostrowski2022ethics,cheong2024causal} rather than the fairness-related challenges that occur due to the employed ML prediction algorithms.
%
\cite{londono2022fairness} presented a survey on fairness in robot learning. 
\cite{ogunyale2018does} conducted experiments to investigate the impact that a robot's ``skin colour" can have on human perceptions of the robot's behaviour.
\cite{ostrowski2022ethics} explored the idea of fairness via the concept of resource allocation.
%
\cite{chang2021unfair} investigated fairness within the context of human-robot teaming where fairness is quantified in terms of each member's contribution.
\cite{hitron2022ai} investigated the effects of a gender-biased robot and its effect on humans' implicit gender stereotypes.
\cite{richards2023machine} evaluated how security mitigation can result in unfairness in human-robot interaction.
\cite{haring2018ffab} outlined the implications that a robot's design has on the a human's bias to interact socially with the robots.
\cite{alarcon2023differential} investigated the human biases present within human-robot versus human-human trust interactions settings.
\cite{habibian2022encouraging} introduced an optimisation approach for fairer subtask allocation.
\cite{lachemaier2024towards} analysed the human automation bias that arose when participants believe a 
robot’s false judgment. 
%
However, none of the existing works have investigated the problem of \textit{ML bias and fairness} within HRI context.
Our work is the first study investigating bias and fairness in ML for HRI.


%

\subsection{Data Augmentation for Bias Mitigation} 
Bias can be mitigated at the pre-processing, in-processing or post-processing stage
\cite{cheong2021hitchhiker}.
The proposed method falls under the pre-processing data augmentation category which has proven to be effective in mitigating bias \cite{cheong2023counterfactual}.
There is minimal work that focus on mitigating bias for a small dataset setup \cite{schnabel2020debiasing}.
For a small dataset problem, \cite{schnabel2020debiasing} leverages on a small annotated dataset to debias a larger dataset.
This is distinct from our work as it focuses specifically on an item recommendation system. 
Existing research has indicated that re-sampling outperforms reweighting for correcting sampling bias \cite{anresampling}.
Given the above, we propose a simple re-sampling or data augmentation method based on the mixup method proposed in \cite{zhangmixup}.
\emph{Mixup} has proven to be a simple yet highly effective method to address challenges ranging from robustness \cite{hendrycksaugmix}, fairness \cite{chuangfair_mixup} and regularisation \cite{yun2019cutmix}.
As a result, \emph{Mixup} has been frequently used as a benchmark for new data augementation techniques and there are recent works proposing new variations of the original method \cite{chuangfair_mixup}.

\section{Problem Formulation}
We study the problem of model fairness using a machine learning approach, where the goal is to predict a correct outcome $y_i\in Y$ from input $\mathbf{x}_i\in X$ based on the available dataset $D$ for individual $i\in I$.
In our setup, $y_i\in Y$ is thus the outcome where $Y=1$ denotes ``high-PA" (i.e., high positive affect, indicative of higher levels of mental wellbeing) whereas $Y=0$ denotes ``low-PA" (i.e., low positive affect, indicative of lower levels of mental wellbeing).
The fairness measure of a model $M$ is then evaluated according to the sensitive groups of individuals defined by their sensitive attributes $A$ gender and race in this work. 
In our experiments, both sensitive attributes analysed are binary. 
They belong to the majority group, e.g.: $A_{race}=1$ if they are White or $A_{race}=0$ if otherwise.
$\hat{Y}$ denotes the predicted class.
%

\subsection{Fairness Measures} 
%
The fairness measures are similar to that in \cite{yan2020mitigating} and \cite{zanna2022bias}.
%
%
%
%
%
\begin{itemize}
    \item \textbf{Equal Accuracy ($EA$)}, a group-based metric, is used to compare the group fairness between the models. 
    This can be understood as the accuracy gap between the majority and the minority group: 
    \begin{equation}
    \label{eqn:equal_accuracy}
    \ EA = |MAE (\hat{Y}| A=1) - MAE (\hat{Y}| A=0)|  ,
    \end{equation}
    where $MAE$ represents the Mean Absolute Error (MAE) of the classification task of each sensitive group.
    \item \textbf{Disparate Impact ($DI$)}, measures the ratio of positive outcome ($\hat{Y}=1$) for both the majority and minority group as represented 
    by the following equation:
    \begin{equation}
    \label{eqn:disparate_impact}
    \ DI = \frac{Pr(\hat{Y}=1|A=0 ) }{ Pr(\hat{Y}=1 | A=1)}
    \end{equation}
%
\end{itemize}
The two measures above represent different aspects of bias. 
$EA$ evaluates fairness based on the model's predictive performance measured in terms of accuracy. 
whereas $DI$ evaluates fairness based purely on the predicted outcomes $\hat{Y}$.


\subsection{Proposed Method: MixFeat}
Our proposed methodology (MixFeat) is based on the data augmentation technique proposed by \cite{zhangmixup}.
Given a dataset of size N where 
$A$ represents the audio cue, 
$F$ represents the facial cue 
and $V$ represents the verbal cue,
the new training sample ($A_k$, $F_k$, $V_k$) is therefore generated as follow:
\begin{equation}
\label{eqn:mixmod}
    \begin{aligned}
    A_k = \lambda_A \cdot A_i + (1-\lambda_A)\cdot A_j\\
    F_k =  \lambda_F \cdot F_i + (1-\lambda_F)\cdot F_j\\
    V_k =  \lambda_V \cdot V_i + (1-\lambda_V)\cdot V_j\\
    \end{aligned}
\end{equation}
where $i,j\in \{1,...N\}$, $i\neq j$ and $\lambda_A , \lambda_F, \lambda_V \sim $ Beta(0,1). 
We use the above method to generate synthetic samples for the minority group to obtain balanced samples across the sensitive attributes of race and gender.
The intuition behind this method is that if we generate new samples by mixing up features from other samples with the same sensitive attribute, the new samples will inherit the sensitive-attribute specific features.
Thus, this method preserves the relation between the synthetic samples and supervision signal which gives the algorithm more samples to learn from without imposing strong assumptions \cite{zhangmixup}.
Figure \ref{fig:model} outlines the experimental setup and how the method is integrated into the overall classification pipeline.

\section{Datasets and Methods}

\subsection{Datasets}

\subsubsection{The AFAR BSFT Dataset}

We collected a dataset of human-human dyadic interactions between a human wellbeing coach and 11 participants over four weeks. The human wellbeing coach was instructed to deliver a Brief-Solution Focused Therapy (BSFT) style coaching, asking participants to focus on solutions rather than analysing the problem \cite{de2012more} for about 20 minutes. After each session, we asked participants to complete the Positive And Negative Affect Scale (PANAS) \cite{watson1988development} to evaluate their positive and negative affect.

\paragraph{Data Collection}
\begin{figure}
    \centering
    \includegraphics[width = 0.9\columnwidth]{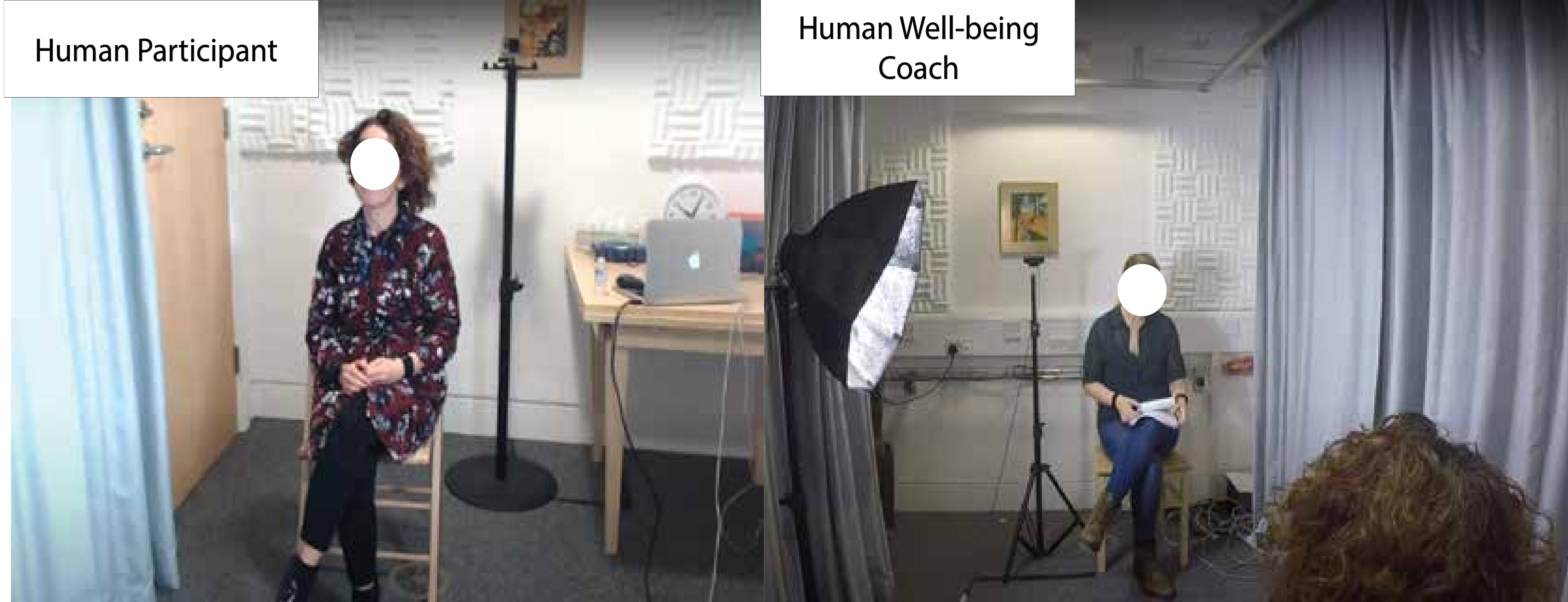}
    \caption{AFAR-BSFT Dataset: interaction between the participants (on the left side), and the human wellbeing coach (on the right side).}
    \label{fig:bsft}
    \vspace{-0.5cm}
\end{figure}

\begin{figure}
    \centering
    \includegraphics[width =0.9\columnwidth]{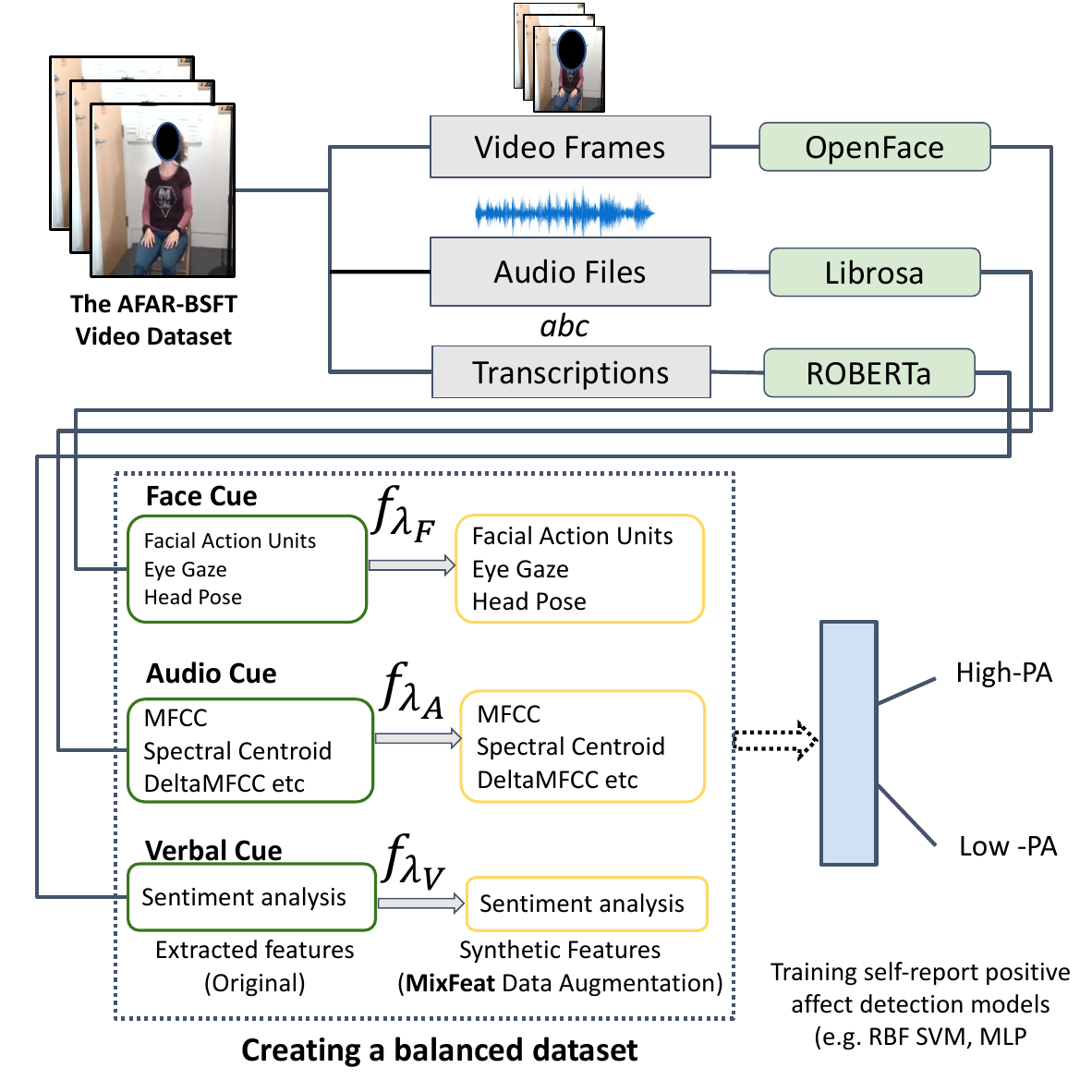}
    \caption{The model pipeline with our proposed data augmentation technique: \textbf{MixFeat}. After extracting the high-level features from the dataset, we generate synthetic sample features using Equation \ref{eqn:mixmod}. Each modality's feature generation process is chiefly governed by their respective $\lambda \sim $ Beta(0,1) parameters.
    }
    \label{fig:model}
    \vspace{-0.5cm}
\end{figure}

11 participants were recruited via email advertising of the University of Cambridge.  We conducted the study in a dedicated room (see Figure \ref{fig:model}) where a human wellbeing coach and one participant were seated in front of each other. Video recordings were done using two external cameras, one facing the participant and the other facing the human coach that can be used for further analysis (beyond the scope of this paper) on dyadic interactions  during the coaching practice. 
We collected 44 videos ($11$ participants $\times$ $4$ weeks, ~20 mins per session) of dyadic wellbeing coaching interactions. 3 out of 44 sessions were excluded due to technical issues (e.g., corrupted video or audio recordings).

\paragraph{Sensitive Groups}
\label{sec:tests}
Two human annotators labelled the gender and race of the participants (with a 100\% agreement). This resulted in 7 participants being labelled as males and 4 as females, and 8 participants being labelled as Whites, and 3 as non-Whites.


\subsubsection{The AFAR Robocoaching 2022 Dataset (AFAR-RC22)}
\label{sec:first-ds}

In 2022, we collected a dataset of human-robot interactions between a robotic mental wellbeing coach and 26 participants over four weeks in a tech company (Cambridge Consultants Inc.) \cite{spitale2023robotic}. The robotic wellbeing coach delivered four positive psychology exercises once a week -- savouring, gratitude, accomplishments, and optimism about the future -- that lasted around 10 minutes each. As we did for the AFAR BSFT dataset, among other measures, we asked participants to fill out the PANAS questionnaire after each interaction session with the robotic coach. The robotic coach was pre-programmed to conduct the positive psychology practice following predefined steps regardless the employees speech. For example, when the robotic coach asked the employee to share what they have been grateful for during the last week, the robot asked the same follow up question to all employees without adapting the coaching to what has been said by the employees.

\begin{figure}
    \centering
    \includegraphics[width = \columnwidth]{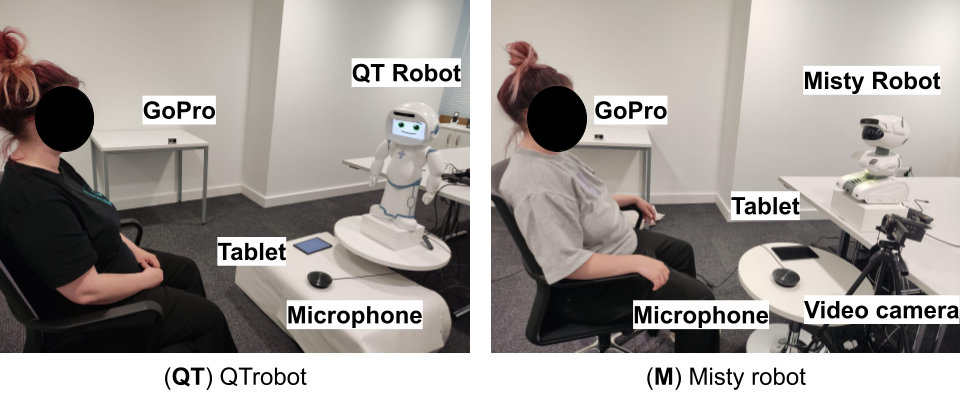}
    \caption{AFAR Robocoaching 2022 Dataset (AFAR-RC22): the setting of the study. (\textbf{QT}) 14 participants interacted with the QT robot; (\textbf{M}) 12 participants interacted with the Misty II robot \cite{spitale2023robotic}.}
    \label{fig:study1}
    \vspace{-0.5cm}
\end{figure}

\paragraph{Data Collection}
Cambridge Consultants Inc advertised the study via their communication channels and the participation was on voluntary-basis. 26 participants that took part were healthy employees of the company. Please refer to our paper \cite{spitale2023robotic} for more information about this study and the recruitment and screening process.
Employees interacted once a week with a robotic coach that delivered positive psychology exercises over four weeks. This was a between-subject study in which employees were randomly assigned to interact either with a QT robot (humanoid-like appearance) or a Misty II robot (toy-like appearance), as shown in Figure \ref{fig:study1}. 
Video recordings were done using an external camera that captured the employees' behaviours during the robotic wellbeing coaching \cite{spitale2023longitudinal}. 
We collected a total of 104 videos (26 participants X 4 weeks, 10 mins each) of robotic wellbeing coaching sessions. 4 of them were excluded due to technical issues (e.g., corrupted audio-visual recordings). 

\paragraph{Sensitive Groups}
Participants self-reported their gender before the study: 6 participants self-reported as female, 1 as non-binary person, and 19 as males. 
While for race, two human annotators labelled the race of participants (with again 100\% agreement), that resulted in 
21 participants labeled as Whites, and 5 participants as non-Whites.

\subsubsection{The AFAR Robcoaching 2023 Dataset (AFAR-RC23)}
We collected the second dataset on longitudinal robot coaching in 2023 as reported in \cite{axelsson2024ohsorry, spitale2023vita} in the same tech company. This dataset collated data recorded into two study. The first study reported in \cite{axelsson2024ohsorry} involved 12 participants -- who had already interacted with a robotic coach in \cite{spitale2023robotic}. 
The second study involved 17 new participants who had never interacted with a robotic coach before. 

\begin{figure}
    \centering
    \includegraphics[width = 0.8\columnwidth]{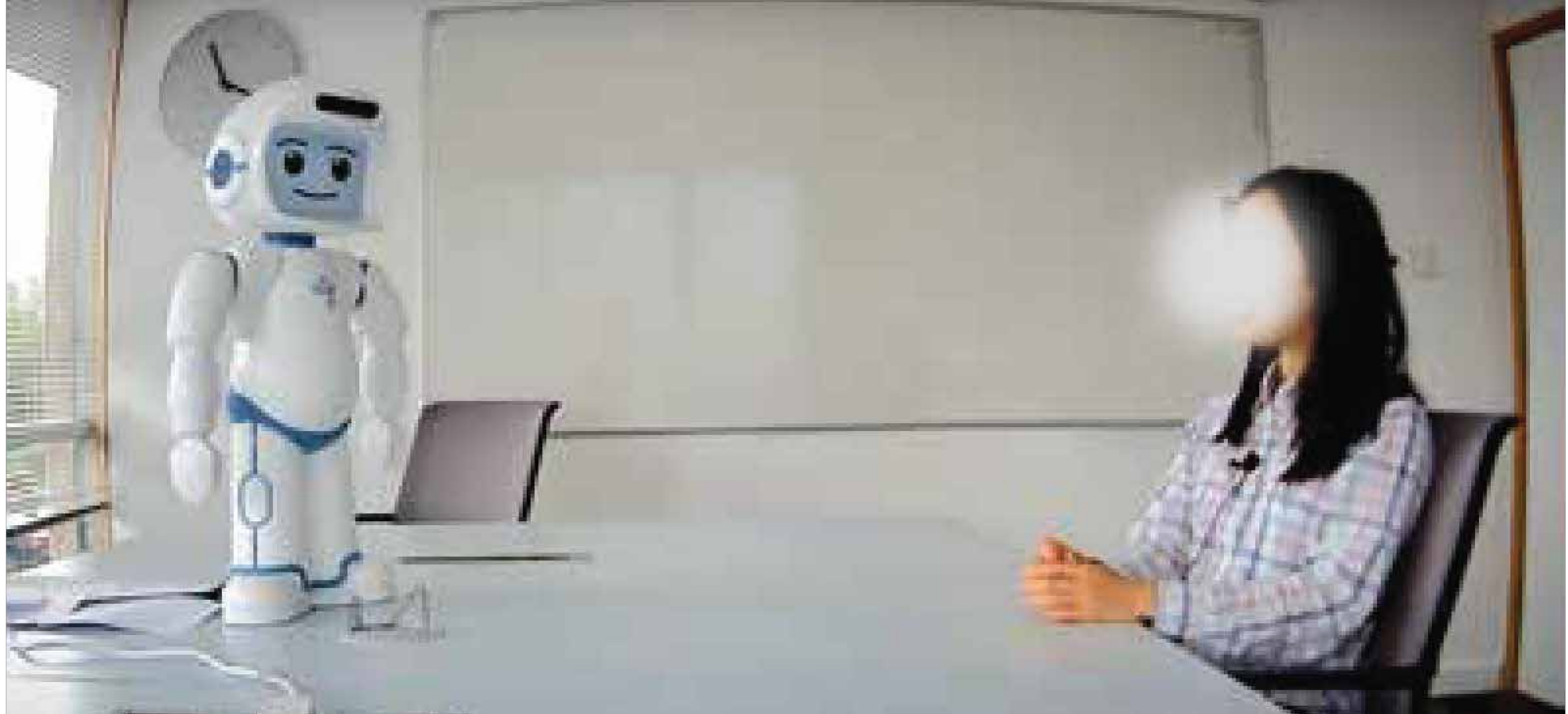}
    \caption{AFAR-Robo Coaching 2023 Dataset (ARC-2023): the setting of the study. 29 participants interacted with the LLM-powered QT robot \cite{spitale2023vita}.}
    \label{fig:study2}
    \vspace{-0.5cm}
\end{figure}

\paragraph{Data Collection}
A total of 29 participants were involved in the two studies with the same recruitment process reported in Section \ref{sec:first-ds}. Please refer to the following papers for more details about the two studies \cite{axelsson2024ohsorry, spitale2023vita}. Employees interacted once a week with the QT robot (the robotic platform chosen for these two studies, see Figure \ref{fig:study2}) that delivered four positive psychology exercises over four weeks -- savouring, gratitude, accomplishments, and one door closes one door opens. Audio-visual data were collected via an external camera that captured the face and body of the employees interacting the the robotic wellbeing coach. We collected a total of 116 videos (29 participants x 4 week, 10 mins for each session). 1 of them was excluded due to technical issues.

\paragraph{Sensitive Groups}
Participants self-reported their gender: 
(study 1) 3 females, 1 non-binary, and 8 males; and (study 2) 7 females, and 10 males. While for the race, two human annotators have labelled the dataset with a full agreement as follows: (study 1) 3 non-Whites and 9 Whites; (study 2) 1 non-White and 16 Whites.

\subsubsection{Annotations}

We assessed the participants' positive affect using the self-report results of the PANAS questionnaire \cite{watson1988development} for all three datasets, which has been widely used by practitioners to identify strengths and concerns in mental wellbeing. 
We computed the positive affect (PA) and negative affect (NA) sub-scales according to the manual in \cite{watson1988development}. 
We set the threshold value to 33.3, corresponding to the mean value for the American population \cite{watson1988development}, and we then classified the videos collected into ``high-PA" and ``low-PA". 
This resulted in: 
(1) AFAR-BSFT DB: 17 videos for the ``low-PA" and 26 videos for the ``high-PA" class; 
(2) AFAR-RC22 DB: 45 videos for the ``low-PA" and 57 videos for the ``high-PA" class; and 
(3) AFAR-RC23 DB: 51 videos for the ``low-PA" and 64 videos for the ``high-PA" class.
Given the small size of all three datasets, we decided to limit our problem to a binary classification problem.

\subsection{Self-report Affect Detection}

\subsubsection{Dataset Pre-processing}
Before extracting the features, we split the audio and video recordings of the three datasets. We asked a human annotator to transcribe the dyadic interactions between the human coach and the participants manually for the AFAR-BSFT DB. The annotator also took note of the timestamp of the speech so that we were able to diarize the audio files. For the AFAR-RC22 dataset, we did not transcribe coachee interactions during robotic coaching. However, for the AFAR-RC23 dataset, we automatically transcribed the coachees' dialogue during interactions with the robotic coach, enabling adaptive conversation responses.

\subsubsection{Multi-modal Feature Extraction}
We extracted the facial features using OpenFace $2.0$ \cite{baltrusaitis2018openface} -- which represents one of the state-of-the-art tools for extracting facial features within the ACII community, e.g., in \cite{mathur2021modeling} -- resulting in the following: eye gaze directions, the intensity and presence of 17 facial action units (FAUs), facial landmarks, head pose coordinates, and point-distribution model (PDM) parameters for facial location, scale, rotation and deformation, resulting in 709 facial features. We adopted this facial features extraction strategy for all three datasets.
We used librosa\footnote{\url{https://librosa.org/doc/latest/index.html}} to extract the audio features for the AFAR-BSFT dataset, namely pitch, speech duration, 128 Mel spectrograms, 20 MFCC, 20 delta MFCC, spectral centroid, and RMS, which results in 172 audio features, as in previous works, e.g., \cite{benssassi2021investigating}. 
While for AFAR-RC22 and AFAR-RC23 DBs, we used openSMILE\footnote{\url{https://www.audeering.com/research/opensmile/}} to extract audio features using the GeMaPs method, that includes e.g., loudness, alpha ratio, hammarberg index, slop, spectral flux and MFCC, that resulted in a total of 25 features, because we have already extracted such features for our previous works \cite{spitale2023robotic, spitale2023vita, axelsson2024ohsorry}.
We used ROBERTa\footnote{\url{https://huggingface.co/docs/transformers/model_doc/roberta}}  
to extract the predicted sentiment for all three datasets from the participants' transcriptions resulting in 2 verbal features (label and probability), as in \cite{abbasi2023computational}.

\subsubsection{Pre-processing}
We first removed constant and null features to prepare the multi-modal features for the machine learning models. Then, we decided to condense the temporal information of each video clip into statistical descriptors as in \cite{mathur2021modeling, abbasi2023computational}, computing a fixed-length vector for each multi-modal feature of each clip that consists of mean, median, standard deviation, minimum, maximum, and auto-correlation with 1-second lag, resulting in: 
(1) AFAR-BSFT DB: a facial feature vector with size $41\times709\times6$, in an audio feature vector with size $41\times 172\times6$, and in a verbal feature vector with size $41\times 2\times6$; 
(2) AFAR-RC22 DB: a facial feature vector with size $100\times709\times4$, in an audio feature vector with size $100\times75\times 4$, and in a verbal feature vector with size $100\times2\times4$; and 
(3) AFAR-RC23 DB: a facial feature vector with size $115\times709\times4$, in an audio feature vector with size $115\times75\times4$, and in a verbal feature vector with size $115\times2\times4$.

\begin{table}[ht]
  \caption{Uni-modal High vs Low-level Feature Modeling Results for the \textbf{AFAR-BSFT DATASET}. R: RBF SVM. M: MLP. Values in bold denote the best outcome
  across experiments.}
  \label{tab: bsft_high_low_uni}
  \centering
  \begin{tabular}{l|cc|cc|cc|cc|cc|cc|}
    \toprule
    &\multicolumn{6}{c|}{Uni-modal}\\
    \midrule  
    &\multicolumn{2}{c|}{Face} &\multicolumn{2}{c|}{Audio}  &\multicolumn{2}{c|}{Verbal}\\
    \midrule   
     &Low  &High &Low  &High &Low  &High  \\
    \midrule    
    R-Acc  &0.27	&\textbf{0.45}&  0.27&\textbf{0.65} & N/A&0.50\\
    R- F1 &0.33&\textbf{0.44}   &0.33&\textbf{0.68}&N/A &0.43\\ 
    M-Acc  &0.35&\textbf{0.43}  &0.35& \textbf{0.47}& N/A&0.49\\
    M- F1 &0.37& \textbf{0.37}&0.37& \textbf{0.37}& N/A& 0.78\\
  \end{tabular}

%
\label{tab: bsft_high_low_multi}
  \centering
  \begin{tabular}{l|cc|cc|cc|}
    \toprule
    &\multicolumn{6}{c|}{Face and Audio} \\
    \cmidrule(r){2-7}
     &\multicolumn{2}{c|}{Early} &\multicolumn{2}{c|}{Soft Voting}  &\multicolumn{2}{c|}{Stacking}\\    
   \cmidrule(r){2-7}
    &Low  &High &Low  &High &Low  &High  \\       
    \midrule   
    R-Acc      & \textbf{0.46}&	0.38& 0.45&\textbf{0.59}	&\textbf{0.76} & 0.71\\
    R-F1        &\textbf{0.53} &	0.48& 0.46&\textbf{0.62}	&\textbf{0.80} &0.76	 \\
    M-Acc   &\textbf{0.43} &	0.33& 0.40&\textbf{0.56}	& 0.71 & \textbf{0.75}\\
    M-F1          &\textbf{0.44} &0.40 &0.52 &\textbf{0.65} & 0.76 & \textbf{0.78}\\    
    \bottomrule
  \end{tabular}
  \vspace{-0.2cm}
\end{table}

\begin{table}[ht]
  \caption{Uni-modal High vs Low-level Feature Modeling Results for the \textbf{AFAR-RC22 DATASET}. 
  Values in bold denote
  the best outcome across the experiments.}
  \label{tab: 1lbrw_high_low_uni}
  \centering
  \footnotesize
  \begin{tabular}{l|cc|cc|cc|cc|}
    \toprule
    &\multicolumn{6}{c|}{Uni-modal}\\
    \midrule  
    &\multicolumn{2}{c|}{Face} &\multicolumn{2}{c|}{Audio} \\
    \midrule   
     &Low  &High &Low  &High  \\
    \midrule    
    R-Acc 	&0.56	&\textbf{0.62}	&0.52	&0.52		\\
    R- F1 &0.68	&\textbf{0.69}	&\textbf{0.65}	&0.63\\ 
    M-Acc &0.55	&\textbf{0.65}	&0.57	&\textbf{0.58}\\
    M- F1 &0.62	&\textbf{0.68}	&\textbf{0.65}	&0.64\\
    \bottomrule
  \end{tabular}

 \label{tab: 1lbrw_high_low_multi}

  \centering
  \begin{tabular}{l|cc|cc|cc|}
    &\multicolumn{6}{c|}{Face and Audio} \\
    \cmidrule(r){2-7}
     &\multicolumn{2}{c|}{Early} &\multicolumn{2}{c|}{Soft Voting}  &\multicolumn{2}{c|}{Stacking}\\    
   \cmidrule(r){2-7}
    &Low  &High &Low  &High &Low  &High  \\       
    \midrule   
    R-Acc     &\textbf{0.58}	&0.55	&0.47	&\textbf{0.53}	&0.52	&\textbf{0.54} \\
    R-F1      &0.66	&\textbf{0.71}	&0.52	&\textbf{0.61}	&0.59	&\textbf{0.62}  \\
    M-Acc     &\textbf{0.59}	&0.52	&0.55	&0.55	&\textbf{0.59}	&0.55\\
    M-F1      &\textbf{0.65}	&0.56	&0.60	&\textbf{0.71}	&0.60	&\textbf{0.71}   \\    
    \bottomrule
  \end{tabular}
  \vspace{-0.2cm}
\end{table}

\begin{table}[ht]
  \caption{Uni-modal High vs Low-level Feature Modeling Results for the \textbf{AFAR-RC23 DATASET}. R: RBF SVM. M: MLP. Values in bold denote the best outcome across the experiments.}
  \label{tab: 2lbrw_high_low_uni}
  \centering
  \begin{tabular}{l|cc|cc|cc|cc|cc|cc|}
    \toprule
    &\multicolumn{6}{c|}{Uni-modal}\\
    \midrule  
    &\multicolumn{2}{c|}{Face} &\multicolumn{2}{c|}{Audio}  &\multicolumn{2}{c|}{Verbal}\\
    \midrule   
     &Low  &High &Low  &High &Low  &High  \\
    \midrule    
    R-Acc &0.63	&\textbf{0.70}	&\textbf{0.55}	&0.54	&\textbf{0.57}	&0.56 \\
    R- F1 &0.74	&\textbf{0.78}	&\textbf{0.66}	&0.65	&0.68	&0.68\\ 
    M-Acc  &0.66	&\textbf{0.73}	&0.51	&\textbf{0.53}	&0.60	&0.60\\
    M- F1 &0.75	&\textbf{0.78}	&0.61	&\textbf{0.69}	&0.69	&0.69 \\
  \end{tabular}

  \label{tab: 2lbrw_high_low_multi}
  \centering
  \begin{tabular}{l|cc|cc|cc|}
    \toprule
    &\multicolumn{6}{c|}{Face and Audio} \\
    \cmidrule(r){2-7}
     &\multicolumn{2}{c|}{Early} &\multicolumn{2}{c|}{Soft Voting}  &\multicolumn{2}{c|}{Stacking}\\    
   \cmidrule(r){2-7}
    &Low  &High &Low  &High &Low  &High  \\       
    \midrule   
    R-Acc  &0.59 &\textbf{0.60}	&0.65 &\textbf{0.72} &0.69	&\textbf{0.72} \\
    R-F1    &0.74 &\textbf{0.75} &0.77	&\textbf{0.78} &0.77 &\textbf{0.78}   \\
    M-Acc &\textbf{0.59} &0.50	&0.63 &\textbf{0.73} &0.62 &\textbf{0.73}\\
    M-F1 &\textbf{0.70}	&0.62 &0.74	&\textbf{0.80}	&0.73 &\textbf{0.79}   \\    
    \bottomrule
  \end{tabular}
  \vspace{-0.2cm}
\end{table}

\subsubsection{Feature Selection}
We defined the high-level and low-level features as interpretable (e.g., facial action unit, pitch) and not-interpretable (e.g., spectral features) to select the most informative ones for the 
positive affect detection model \cite{du2019techniques}. The low-level features were 1) facial: facial landmarks, head pose coordinates, and point-distribution model (PDM) parameters, and 2) audio: 128 Mel spectrograms, 20 MFCC, 20 delta MFCC, spectral centroid, and RMS for AFAR-BSFT dataset and MFCC, shimmer, jitter, F0, F1, F2, and F3 as audio features for the AFAR-RC22 and 23 DBs; while the high-level features were 1) facial: facial action units and gaze, 2) audio: pitch and speech duration for AFAR-BSFT dataset and loudness, alpha ratio, hammarberg index, slop, spectral flux for the AFAR-Robo Coaching 2022 and 2023 DBs, and 3) verbal: the sentiment of the speech. 
Given the differences in dimensionality between low-level and high-level features, we conducted a principal component analysis (PCA) to reduce the size of the features while keeping 80\% of the information. 
The PCA analysis resulted in: 

\begin{itemize}

    \item \textbf{AFAR-BSFT.} 
    i) 5 principal components (PCs) for high-level features for face, 10 PCs for low-level features for face, and 
    ii) 2 features for high-level features for audio (no PCA conducted because the number of high-level audio features was already small, i.e., equal to 2), and 3 PCs for low-level features for audio.
    \item \textbf{AFAR-RC22 DB:} 
    i) 19 PCs for high-level features for face, 5 PCs for low-level features for face, and 
    ii) 4 PCs for high-level features for audio, and 3 PCs for low-level features for audio.
    \item \textbf{AFAR-RC23):} 
    i) 17 PCs for high-level features for face, 6 PCs for low-level features for face, and 
    ii) 3 PCs for high-level features for audio, and 6 PCs for low-level features for audio.
\end{itemize}

\subsubsection{Data Fusion Strategies}
We explored different state-of-the-art data fusion strategies \cite{atrey2010multimodal} for all three datasets. We experimented with early fusion, which consisted of concatenating features from different modalities that resulted in a single vector of features, 
and different late fusion strategies, namely  majority voting (soft and hard) and stacking (soft and hard). In majority voting, the final decision is made according to the most frequent class label predicted across the different uni-modal models (hard) or the classifier whose predicted class probability is the highest across the different uni-modal models (soft). In stacking, the final decision is made by another classifier (e.g., logistic regression model) fed by either the predicted class label (hard) or the predicted class probabilities (soft) of each uni-modal model.



\begin{table*}[ht]
  \caption{Unimodal and Multimodal Debiasing Results for the \textbf{AFAR-BSFT DATASET}. Abbreviations. R: RBF SVM. M:MLP. UAR: Unweighted Average Recall. Values in bold denote the best outcome across the three sets of experiments.}
  \label{tab: bsft_uni_results}
  \centering
   \footnotesize
    \addtolength{\tabcolsep}{-0.5mm}
  \begin{tabular}{l|cc|cc|cc|cc|cc|cc|cc|cc|cc|}
    \toprule
    &\multicolumn{6}{c|}{Original} &\multicolumn{6}{c|}{Baseline Comparison} &\multicolumn{6}{c|}{Proposed Method}\\
    \cmidrule(r){2-19}
    &\multicolumn{2}{c|}{Face} &\multicolumn{2}{c|}{Audio}  &\multicolumn{2}{c|}{Verbal}
    &\multicolumn{2}{c|}{Face} &\multicolumn{2}{c|}{Audio}  &\multicolumn{2}{c|}{Verbal}
    &\multicolumn{2}{c|}{Face} &\multicolumn{2}{c|}{Audio}  &\multicolumn{2}{c|}{Verbal}\\
    \cmidrule(r){2-19}
    
     &R  &M &R &M &R &M &R  &M &R &M &R &M  &R  &M &R &M &R &M  \\
    
    \midrule
    \textbf{Overall Acc}  &0.41	&0.46 & 0.54	&0.34 &0.34	&0.46 &0.43	&0.40 &0.64	&0.41 &0.29	&0.59 &\textbf{0.59}	&0.43 &\textbf{0.81}	&0.55 &0.45	&\textbf{0.62} \\
    \textbf{Overall F1} & 0.57&		0.52&		0.68&		0.37&		0.43&		0.78& 0.51	&	0.43&		0.77&		0.56	&	0.29&		0.73& 0.72		&0.50&		0.83	&	0.68&		0.52	&	0.70\\
    \textbf{Overall UAR} &0.42	&0.39	&0.58	&0.35	&0.33	&0.52	&0.43	&0.40	&0.64	&0.41	&0.29	&0.59	&0.59	&0.43	&0.81	&0.55	&0.45	&0.62\\

    \midrule
    $EA_{Gender}$ &\textbf{0.02}	&0.49 &0.27	&\textbf{0.02} &0.08	&0.38 &0.03	&0.45 &0.38	&0.21 &0.10	&0.48 &0.07	&0.38 &0.31	&0.14 &0.07	&0.21\\
    $EA_{Race}$   &0.23	&\textbf{0.07} &0.05	&0.11 &0.36	&0.07 &0.10	&0.17 &0.17	&0.21 &\textbf{0.07}	&0.21 &\textbf{0.07}	&0.10 &0.10	&\textbf{0.01} &0.17	&0.14\\
    $DI_{Gender}$ &\textbf{0.88}	&0.72 &1.29	&1.74 &0.91	&1.23 &0.75	&0.63 &1.26	&1.82 &0.67	&1.21 &0.82	&0.75 &1.21	&\textbf{1.19} &\textbf{0.94}	&1.37\\
    $DI_{Race}$  &0.68 &0.12 &\textbf{0.97}	&1.41 &0.60	&1.06 &0.68	&0.24 &1.08	&1.58 &0.50 &1.12 &0.76	&0.33 &1.33	&1.06 &0.84	&0.96\\
    \bottomrule
  \end{tabular}


  \label{tab: bsft_multi_results}
  \centering
  
  \vspace{1mm}
  \begin{tabular}{l|cc|cc|cc|cc|cc|cc|cc|cc|cc|}
     &\multicolumn{2}{c|}{Early} &\multicolumn{2}{c|}{Soft Voting}  &\multicolumn{2}{c|}{Stacking}
    &\multicolumn{2}{c|}{Early} &\multicolumn{2}{c|}{Soft Voting}  &\multicolumn{2}{c|}{Stacking}
    &\multicolumn{2}{c|}{Early} &\multicolumn{2}{c|}{Soft Voting}  &\multicolumn{2}{c|}{Stacking}\\
    
   \cmidrule(r){2-19}
    &R  &M &R &M &R &M &R  &M &R &M &R &M  &R  &M &R &M &R &M  \\    
   
    
    
    \midrule
    \textbf{Overall Acc} &0.54	&0.51 &0.66	&0.44 &0.54	&0.51 &0.69	&0.48 &\textbf{0.71}	&0.52 &\textbf{0.69}	&0.57 &\textbf{0.72}	&0.62 &0.69	&0.55 &0.67	&0.59  \\
    \textbf{Overall F1}&0.67&		0.55	&	0.74	&	0.51&		0.60		&0.63 & 0.75	&	0.48&		0.71&		0.63&		0.68&		0.74& 0.71	&	0.72&		0.69&		0.61&		0.64&		0.56\\
    \textbf{Overall UAR} &0.52	&0.42	&0.67	&0.44	&0.53	&0.58	&0.69	&0.48	&0.71	&0.52	&0.57	&0.69	&0.72	&0.62	&0.69 &0.55	&0.67	&0.59\\
    
    \midrule
    $EA_{Gender}$ &\textbf{0.05}	&0.56 &0.08	&\textbf{0.02} &\textbf{0.06}	&0.42 &0.14	&0.28 &0.03	&0.14 &0.17	&0.28 &0.07	&0.07 &0.07	&0.14 &0.11	&0.07  \\
    $EA_{Race}$   &0.17	&\textbf{0.14} &0.22	&\textbf{0.03} &0.29	&0.14 &\textbf{0.14}	&\textbf{0.14 }&0.31	&0.07 &0.28	&0.07 &0.28	&0.21 &0.41	&0.07 &0.21	&0.14     \\
    $DI_{Gender}$  &1.10 &0.83 &1.69 &1.34 &1.24 &1.47 &0.76 &0.81 &\textbf{0.88} &1.20 &1.57 &1.44 &\textbf{0.96} &1.06 &1.15 &0.84 &\textbf{0.90} &1.29 \\
    $DI_{Race}$   &0.91	&0.13 &1.21	&1.38 &0.85	&1.40 &0.68	&0.26 &0.78	&\textbf{0.94} &1.40	&\textbf{0.95} &\textbf{1.04}	&0.68 &1.26	&\textbf{0.94} &0.81	&\textbf{0.95 } \\
    \bottomrule
  \end{tabular}
\end{table*}


\begin{table*}[ht]
  \caption{Unimodal and Multimodal Debiasing Results for the \textbf{AFAR-RC22 DATASET}. Abbreviations. R: RBF SVM. M:MLP. UAR: Unweighted Average Recall. Values in bold denote 
  the best outcome across the three sets of experiments.}
  \label{tab: 1lbrw_uni_results}
  \centering
   \footnotesize
\addtolength{\tabcolsep}{-0.5mm}
  \begin{tabular}{l|cc|cc|cc|cc|cc|cc|cc|}
    \toprule
    &\multicolumn{4}{c|}{Original} &\multicolumn{4}{c|}{Baseline Comparison} &\multicolumn{4}{c|}{Proposed Method}\\
    \cmidrule(r){2-15}
    &\multicolumn{2}{c|}{Face} &\multicolumn{2}{c|}{Audio} 
    &\multicolumn{2}{c|}{Face} &\multicolumn{2}{c|}{Audio}  
    &\multicolumn{2}{c|}{Face} &\multicolumn{2}{c|}{Audio} \\
    \cmidrule(r){2-15}
    
     &R  &M &R &M &R &M &R  &M &R &M &R &M   \\
    
    \midrule
    \textbf{Overall Acc} &0.39	&0.54	&0.48	&0.52 &0.56	&0.51	&0.54	&0.56 &\textbf{0.59} &0.55 &\textbf{0.59}	&\textbf{0.59} \\
    \textbf{Overall F1} &0.55 &0.60	 &0.58	&0.59 &0.66	&0.57	&0.58	&0.51 &\textbf{0.68} &0.65	&\textbf{0.68}	&0.67\\
    \textbf{Overall UAR} &0.57	&0.61	&0.59	&0.60 &0.69	&0.58	&0.59	&0.52 &\textbf{0.70} &0.67	&\textbf{0.70}	&0.68\\

    \midrule
    $EA_{Gender}$ &\textbf{0.01}	&0.10 &0.13	&0.12 &0.08	&0.06 &0.08  &0.08 &0.02 &0.09 &0.18 &0.13\\
    $EA_{Race}$ &0.04 &0.05	&\textbf{0.01} &0.07 &0.22 &0.21 &0.17   &0.17 &0.17 &0.23 &0.26		&0.19\\
    $DI_{Gender}$	&1.02 &1.11 &\textbf{1.00} &0.99  &1.08	&0.75 &0.82	&0.73 &1.01	&0.77 &0.93	&0.78\\
    $DI_{Race}$ &0.83 &0.84	&1.09 &1.45 &0.91 &1.13 &1.11 &1.24 &0.97 &1.10	&\textbf{0.99} &1.20\\
    \bottomrule
  \end{tabular}


  \label{tab: 1lbrw_multi_results}
  \vspace{1mm}
  \centering
  \begin{tabular}{l|cc|cc|cc|cc|cc|cc|cc|cc|cc|}
    \cmidrule(r){2-19}
     &\multicolumn{2}{c|}{Early} &\multicolumn{2}{c|}{Soft Voting}  &\multicolumn{2}{c|}{Stacking}
    &\multicolumn{2}{c|}{Early} &\multicolumn{2}{c|}{Soft Voting}  &\multicolumn{2}{c|}{Stacking}
    &\multicolumn{2}{c|}{Early} &\multicolumn{2}{c|}{Soft Voting}  &\multicolumn{2}{c|}{Stacking}\\
    
   \cmidrule(r){2-19}
    &R  &M &R &M &R &M &R  &M &R &M &R &M  &R  &M &R &M &R &M  \\    
   
    
    
    \midrule
    \textbf{Overall Acc} &0.54	&0.50	&0.62	&0.57	&0.55	&0.60  &0.61 &0.53 &0.59	&0.56	&0.63 &0.63 &\textbf{0.66}	&0.54	&0.60	&0.61	&\textbf{0.68}	&0.61\\
    \textbf{Overall F1} &0.69	&0.55	&0.64	&0.63	&0.62	&0.57 &0.70	 &0.64 &0.68	&0.65	&0.68	&0.71 &\textbf{0.80} &0.69	&0.83	&\textbf{0.84}	&0.77	&\textbf{0.78} \\
    \textbf{Overall UAR} &0.75	&0.55	&0.64	&0.63	&0.62	&0.59 &0.73	&0.66	&0.70	&0.67	&0.69	&0.74 &\textbf{0.78}	&0.69	&0.83	&\textbf{0.84}	&0.77	&\textbf{0.82}\\
    
    \midrule
    $EA_{Gender}$ &0.08	 &0.07 &0.08 &0.09	&\textbf{0.00} &0.17	&0.08 &0.02	&0.11 &0.08	&0.14 &0.02  &\textbf{0.01}  &0.03  &0.13  &\textbf{0.01}  &0.25  &0.06\\
    $EA_{Race}$  &0.24	&0.10 &\textbf{0.05}	&0.28 &0.04	&0.11 &0.22	&0.17 &0.19	&0.17 &0.11	&0.05  &0.13 &\textbf{0.03}	 &0.17	 &\textbf{0.05} &0.17  &\textbf{0.01}\\
    $DI_{Gender}$ &0.93	&1.21 &\textbf{1.04}	&0.83 &1.34	&1.37 &\textbf{0.94}	&0.71 &0.85 &0.73 &\textbf{0.98} &0.87 &0.93 &0.85 &0.88 &0.77 &1.18 &0.83\\
    $DI_{Race}$ &1.07 &0.90 &\textbf{1.03} &1.37	&0.81 &1.09 &1.04 &1.19	&1.07 &1.24	&1.09 &1.21 &1.09 &\textbf{1.04} &1.11 &1.26	&\textbf{1.01} &1.27\\
    \bottomrule
  \end{tabular}
\end{table*}


\begin{table*}[ht]
  \caption{Unimodal and Multimodal Debiasing Results for the \textbf{AFAR-RC23 DATASET}. Abbreviations. R: RBF SVM. M:MLP. UAR: Unweighted Average Recall. Values in bold denote best outcome 
  across the three sets of experiments.}
  \label{tab: 2lbrw_uni_results}
  \centering
  \footnotesize
\addtolength{\tabcolsep}{-0.5mm}
  \begin{tabular}{l|cc|cc|cc|cc|cc|cc|cc|cc|cc|}
    \toprule
    &\multicolumn{6}{c|}{Original} &\multicolumn{6}{c|}{Baseline Comparison} &\multicolumn{6}{c|}{Proposed Method}\\
    \cmidrule(r){2-19}
    &\multicolumn{2}{c|}{Face} &\multicolumn{2}{c|}{Audio}  &\multicolumn{2}{c|}{Verbal}
    &\multicolumn{2}{c|}{Face} &\multicolumn{2}{c|}{Audio}  &\multicolumn{2}{c|}{Verbal}
    &\multicolumn{2}{c|}{Face} &\multicolumn{2}{c|}{Audio}  &\multicolumn{2}{c|}{Verbal}\\
    \cmidrule(r){2-19}
    
     &R  &M &R &M &R &M &R  &M &R &M &R &M  &R  &M &R &M &R &M  \\
    
    \midrule
    \textbf{Overall Acc} &0.46	&0.55 &0.48 &0.46	&0.53 &0.55 &0.57 &0.49 &0.54 &0.57 &0.54	&0.57 &\textbf{0.63}	&0.57	&0.57 &\textbf{0.59}	&0.59	&\textbf{0.63} \\
    \textbf{Overall F1} &0.52 &0.49	&0.49 &0.45	&0.53 &0.60 &0.66 &0.55	&0.54 &0.49 &0.54 &0.49 &\textbf{0.71}		&0.66		&0.64		&\textbf{0.66}		&\textbf{0.66}		&0.41\\
    \textbf{Overall UAR} &0.55	&0.52	&0.50 &0.47	&0.55 &0.62 &0.70 &0.56	 &0.55 &0.49	&0.55 &0.49 &\textbf{0.74} &0.68	&0.65	&\textbf{0.67}	&\textbf{0.68}	&0.42\\

    \midrule
    $EA_{Gender}$ &0.16	&0.21 &0.21 &0.13 &0.04 &0.18 &0.11	&\textbf{0.02} &0.06	&0.04 &0.06	&0.04 &0.23	&0.19 &\textbf{0.01}	&0.04	&0.07 &\textbf{0.03}\\
    $EA_{Race}$  &0.05	&\textbf{0.04} &0.07	&\textbf{0.01} &0.07 &0.03 &0.25 &0.17	&0.15 &0.18	&0.15	&0.18 &0.37	&0.27 &0.10	&0.13 &0.05	&0.06\\
    $DI_{Gender}$ &1.34	&1.41 &1.24	&1.03 &1.22 &0.96 &1.27	&0.79 &1.08	&0.76	&\textbf{1.08}	&0.76 &\textbf{1.14}	&0.84 &\textbf{1.00}	&0.75 &1.22	&0.84\\
    $DI_{Race}$ &\textbf{0.98} &0.90	&1.17 &1.34	&1.31 &1.10 &0.78 &1.07	&0.85	&\textbf{1.10} &0.85 &\textbf{1.10} &0.86	&1.08 &\textbf{0.91}	&1.20 &0.88	&1.17\\ 
    \bottomrule
  \end{tabular}


  \label{tab: 2lbrw_multi_results}
  \centering
  \vspace{1mm}
  \begin{tabular}{l|cc|cc|cc|cc|cc|cc|cc|cc|cc|}
     &\multicolumn{2}{c|}{Early} &\multicolumn{2}{c|}{Soft Voting}  &\multicolumn{2}{c|}{Stacking}
    &\multicolumn{2}{c|}{Early} &\multicolumn{2}{c|}{Soft Voting}  &\multicolumn{2}{c|}{Stacking}
    &\multicolumn{2}{c|}{Early} &\multicolumn{2}{c|}{Soft Voting}  &\multicolumn{2}{c|}{Stacking}\\
    
   \cmidrule(r){2-19}
    &R  &M &R &M &R &M &R  &M &R &M &R &M  &R  &M &R &M &R &M  \\    
   
    
    
    \midrule
    \textbf{Overall Acc} &0.49	&0.42 &0.60 &0.55 &0.57	&0.61 &0.63	&0.54	&0.61	&0.59	&0.62	&0.60 &\textbf{0.67}		&0.62	&0.62	&\textbf{0.70}	&\textbf{0.68}	&0.59\\
    \textbf{Overall F1} &0.61	&0.41 &0.52	&0.61	&0.55	&0.57 &0.70	&0.63	&0.68	&0.66	&0.66	&0.68 &\textbf{0.73}		&0.64	&0.68	&\textbf{0.73}	&\textbf{0.70}	&0.67\\
    \textbf{Overall UAR} &0.63	&0.41 &0.52	&0.61	&0.57	&0.60 &0.73	&0.66	&0.70	&0.68	&0.67	&\textbf{0.70} &\textbf{0.76}		&0.64	&0.71	&\textbf{0.74}	&\textbf{0.70}	&\textbf{0.70}\\
    
    \midrule
    $EA_{Gender}$ &0.05	&0.07 &0.07	&\textbf{0.05} &\textbf{0.02}	&0.16 &0.02	&0.05	&\textbf{0.05}	&0.07	&0.04	&0.04 &\textbf{0.00}	&0.02	&0.06	&\textbf{0.05}	&0.03	&\textbf{0.02}\\
    $EA_{Race}$  &\textbf{0.02}	&0.12 &0.02 &0.31 &0.03 &0.10  &0.14	&0.12	&0.12	&0.14		&0.06	&\textbf{0.03} &0.10	&0.07	&\textbf{0.03}	&0.07	&0.07	&0.05\\
    $DI_{Gender}$ &1.06	&1.52 &\textbf{1.08}	&0.75 &1.48 &1.51 &0.88	 &0.66	&0.79	&0.74	&0.94	&0.87 &\textbf{1.01}		&0.71	&0.89	&0.90	&1.04	&\textbf{0.97}\\
    $DI_{Race}$   &0.78	&0.50	&1.09	&1.37 &0.82	&1.10 &\textbf{1.13}	&1.24	&\textbf{1.07}	&1.25	&1.06	&1.23 &1.13	&1.21 &1.16	&1.14	&\textbf{1.04}	&1.16\\
    \bottomrule
  \end{tabular}
\end{table*}


\section{Modeling and Bias Analysis Results}

\subsection{Modeling and Feature Selection}

We first conducted experiments using various machine learning techniques as in \cite{mathur2021modeling, abbasi2023computational} -- namely logistic regression, linear support vector machine (SVM), random forest tree, bagging, XGBoost, AdaBoost, decision tree, radial basis function support vector machine (RBF-SVM), multi-layer perceptron (MLP), and long-short term memory (LSTM) neural network -- and validating them with three different cross-validation approaches (i.e., 5-fold CV and leave-one-subject-out (LOSO)). Our results showed that the outperforming models were RBF-SVM  and MLP  among the machine learning techniques we experimented with. Due to space constraints, we only report the bestperforming model results and analyses in the following sections.


\subsection{Low vs High Level Feature Analysis}
\subsubsection{AFAR-BSFT}
We trained different experimental models with either the high or low-level features, and compared their performances. Table \ref{tab: bsft_high_low_uni} reports the results of the uni-modal models, while Table \ref{tab: bsft_high_low_multi} reports the results of the multi-modal (i.e., face and audio)
models.
We have not reported the tri-modal (i.e., face, audio, and verbal) analysis because the verbal feature vector contains only high-level information, making comparison impossible.
Our results showed that the models trained with high-level features performed better in terms of accuracy and F1 in all uni-modal and most of the multi-modal setups (see Tables \ref{tab: bsft_high_low_uni} and \ref{tab: bsft_high_low_multi}). 
Therefore in the rest of our work, we only considered high-level features to train the models and conduct the bias analysis.

\subsubsection{AFAR-RC22}
For the AFAR-RC22 dataset, with reference to Table \ref{tab: 1lbrw_high_low_uni}, we see a similar trend where the high-level features are better for the face modality. Across the audio modality, though the low features seem to perform better for F1 scores across both the RBM SVM and MLP methods, the gap in results are minor.  
Across the multimodal setup, with reference to Table \ref{tab: 1lbrw_high_low_multi}, we see a similar trend with the AFAR-BSFT dataset. For the early fusion strategy, low-level features seem to perform better whereas for the late fusion strategies (soft voting and stacking) high-level features performed better.

\subsubsection{AFAR-RC23}
For the AFAR-RC23 dataset, with reference to Table \ref{tab: 2lbrw_high_low_uni}, we see a similar trend where the high-level features are better for the face modality. Across the audio modality, RBF SVM performed better with low level features and MLP performed better for high level features. 
Across the verbal modality, both models seem to produce similar results across both high and low level features. 
Across the multimodal setup, with reference to Table \ref{tab: 2lbrw_high_low_multi} , we see a similar trend with the two other datasets. The key difference is that across early fusion, the RBM SVM performed better using high level features whereas the MLP performed better with low level features. 
For the late fusion strategies (soft voting and stacking), high-level features performed better just as before.

\subsection{Uni-modal vs Multi-modal Analysis}

We conducted several experiments to compare uni-modal and multi-modal (with either early or late fusion) approaches. 

\subsubsection{AFAR-BSFT}
As observed in Tables \ref{tab: bsft_uni_results} and \ref{tab: bsft_multi_results},
overall the multi-modal approach outperformed the uni-modal models. 
Interestingly, the audio modality consistently gives the best accuracy and fairness scores across all three sets of experiments. This could be due to the fact that BSFT coaching is dialogue oriented. 
Across fairness, not all multimodal approaches led to bias reduction. For example, the MLP-based soft major voting approach seems to reduce gender and race biases with respect to audio or face unimodal approaches, however, early fusion techniques for both MLP and RBF SVM-based approaches increase both gender and race biases. We hypothesise that this may be due to fact that early fusion simply combines features without considering their gender-based associations which caused the underlying bias to be amplified by the respective models

\subsubsection{AFAR-RC22}
We see from Tables \ref{tab: 1lbrw_uni_results} and \ref{tab: 1lbrw_multi_results} that overall, a multimodal approach is often better than a unimodal approach across both performance and fairness.
For instance, all of the original multimodal fusion methods (early fusion, soft voting and stacking) gave improvements across accuracy and UAR compared to the best performing unimodal modality and model.
Across fairness, stacking seems to produce the best improvement across the $EA$ metrics whereas soft voting seems to produce the best improvement across the $DI$ metrics.

\subsubsection{AFAR-RC23}

We see that previous observations are generally consistent. We see from Tables \ref{tab: 2lbrw_high_low_uni} and  \ref{tab: 2lbrw_high_low_multi} that a multimodal approach is often better than the best performing unimodal approach across most 
measures. 
The best performing accuracy results improved from $0.57$ to $0.61$ and the best performing UAR results improved from $0.62$ to $0.63$.
Across fairness, 
the best performing original unimodal
$EA_{Gender}$ results improved from $0.04$ to $0.02$. 

\section{Debiasing Approach and Results}

To provide a comparison to our proposed method, we use a baseline data balancing method to mitigate the bias present.
We employ a similar data balancing method as \cite{yan2020mitigating}.
We re-sample the minority group by randomly oversampling datapoints to obtain an augmented dataset with samples balanced across both sensitive attributes.

The implementation of our proposed method is similar to that of the baseline method. 
The key difference is that instead of randomly oversampling data points, we generate synthetic samples according to the method outline in Equation \ref{eqn:mixmod}.

We implemented this data balancing and proposed method similar to all three DBs.
\subsection{AFAR-BSFT}
%

%
After data balancing, we retrain the models and capture the results in Table \ref{tab: bsft_uni_results} and \ref{tab: bsft_multi_results}.
%
%
%

%
%
%
With reference to Table \ref{tab: bsft_uni_results}, we see that across the uni-modal experiments, our method consistently produces a more accurate and fairer outcome across most metrics for both sensitive attributes compared to the baseline. 
%
%
Within the multi-modal approach depicted in Table \ref{tab: bsft_multi_results}, we see that this gap in predictive and fairness performance is diminished. 

\subsection{AFAR-RC22}

%
%
Within a unimodal setting, we see from Table \ref{tab: 1lbrw_uni_results} that our proposed method, MixFeat consistently produces the best results acorss all performance measures for both modalities. 
This is also true within a multimodal setting. 
Across fairness, our results are also better compared to baseline. For instance, across early fusion, our proposed data augmentation method resulted in the fairest $EA_{Gender}$ score of $0.01$ with the RBM SVM classifier and the fairest 
$EA_{Race}$ score of $0.03$ with the MLP classifier.
%
%
%
Within the multi-modal approach in Table \ref{tab: 1lbrw_multi_results}, we see that our proposed method still consistently produces the best results across all performance metrics. 
Across fairness, though our results perform better across most fusion methods 
we see that the baseline methods produces better fairness scores across $DI_{Gender}$ across both the early fusion ($0.94$) and stacking ($0.98$) approaches.

\subsection{AFAR-RC23}
%
We see a consistent trend across the unimodal results in Table \ref{tab: 2lbrw_uni_results}. For instance, for the face modality, we achieved an overall accuracy, F1 and UAR of $0.63$, $0.71$ and $0.74$ respectively. Across the audio modality, our proposed method also produced the best results across $EA_{Gender}$, $DI_{Gender}$ and $DI_{Race}$ with a score of $0.01$, $1.00$ and $0.91$ respectively. 
Across the multimodal results, 
\ref{tab: 2lbrw_multi_results}, our proposed method also consistently produces good results compared to the baseline data augmenation method across both performance and fairness measures. 
%
%
With reference to Table \ref{tab: 2lbrw_uni_results}, we see that both data augmentation methods were effective at improving performance and reducing bias. 
%
%
Within the multi-modal approach depicted in Table \ref{tab: 2lbrw_multi_results}, we see a similar trend. In general, we see our proposed method performing better than the baseline across most fusion strategies. 

\section{Recommendations and Conclusion}
\begin{table*}[]
\caption{Recommendations for Affective and HRI Researchers For Reducing ML Bias in Small Datasets.} 
\footnotesize
\label{tab:recommendations}
\resizebox{\textwidth}{!}{%
\begin{tabular}{llll}
\toprule
\textbf{Research Field}   & \textbf{Recommendations}                                            & \textbf{Why?} & \textbf{How?} \\
\midrule

ML Training& (\textbf{R1}) Train human-centric models                        &  The problem of bias is often             &            Ensure that participants are balanced across    \\
&  on a balanced dataset                  & associated with data imbalances \cite{chakraborty2021bias}.           &          sensitive groups (e.g., gender, race) or       \\
&&& perform data balancing methods\\

&&& as needed (e.g., data augmentation \cite{cheong_acii}).\\
\midrule
ML Features Selection& (\textbf{R2}) Use higher level features making  & The high-level features often include & Extract and train the model using \\
 &     use of multimodality                       &              less noisy information and make &   multimodal high-level features            \\
  &                  &             easier for the model to learn the   &               by experimenting different fusion   \\
    &                       &   representations in small datasets \cite{song2020spectral}  &       strategies \cite{mathur2020introducing}      \\
      &                       &   if more information (i.e., multimodal)  is used.  &              \\
 \midrule
ML Modeling & (\textbf{R3}) Employ a variety of ML models   &  Past works \cite{mathur2020introducing, abbasi2022computational} showed    that & Conduct experiments using a variety\\
& and evaluation and fairness                    &  specific models work better on certain         &   of different models and metrics \cite{mathur2021modeling}.           \\
&  metrics                   &   modalities across specific metrics.         &               \\
\midrule
AC/Robotic System Design & (\textbf{R4}) Balance the trade off between   & Adaptive and more complex AI-based  &Adaptive models that need to be embedded  \\
&AC/robotic system complexity and         &  system may be more difficult             &    in a robot should be tested for their                \\

&   fairness                 &  to debias \cite{yang2022adaptive}.               &        fairness in advance \cite{ma2023fairness}. When this is not     \\
&                    &                &   applicable, on-the-fly model should have          \\
&                    &                &  embedded bias mitigation strategies.       \\




\midrule
AC/Robotic System Design& (\textbf{R5}) Define field and   &    The ethical research in robotics studies the    & Adopt human-centric approach, like value-         \\ 
 \& Ethics &  context specific ethical principles   & consequences of deploying robots in social &  sensitive design \cite{friedman1996value}, to distill main ethical \\ 
&  when designing / deploying AC  &  contexts and of interacting with humans.         &  recommendations to use robots in a specific      \\ 
&     systems for wellbeing                         &   The study and definition of   ethical           &      context \cite{axelsson2022robots}  \\
&                              &    guidelines can help  building fair  HRI   \cite{londono2022fairness}          &             \\
\bottomrule
\end{tabular}%
}
\end{table*}

This paper highlighted the problem of bias in small datasets in the context of human-human and robot-human wellbeing coaching and illustrated that it is possible to debias a small dataset using a principled and methodological approach. 
Efforts must be made to ensure fairness, especially in robot-human interactions, given the responsibility of building and designing these systems in sensitive contexts such as wellbeing.
In addition to our findings reported in \cite{cheong_acii}, we have also noted that as much as it is possible to debias small datasets, the performance of a model, as measured using the standard metrics such as accuracy an F1, is still very correlated and dependent on the size of the dataset. 
AFAR-RC22 (26 participants with a total of 101 datapoints) and AFAR-RC23 (29 participants with a total of 116 datapoints) are bigger than AFAR-BSFT (11 participants with a total of 41 datapoints).
We see from our results that the same models trained on the AFAR-RC22 and AFAR-RC23 datasets perform better across measures such as overall accuracy and overall F1. In addition, 
The effects of data augmentation and data balancing in these dataset produced greater improvements in performance metrics compared to the BSFT dataset. This is likely due to the fact that we had to generate more synthetic samples in order to balance the samples across the different sensitive attribute groups for the larger datasets (AFAR-RC22 and AFAR-RC23). As a result, the increase in training set is likely to have introduced more data for the model to better learn from.
From our findings, we distilled a set of recommendations \textbf{(R1 - R5)} in Table \ref{tab:recommendations} that can be used by AC and HRI researchers to integrate and address fairness-related concerns within their ML-based research  when working with small datasets.\\
\textbf{R1: Train human-centric models on a balanced dataset.} 
Our results suggest that employing an imbalanced human dataset may lead to fairness issues when training ML models. 
Small datasets are commonly utilised in studies within Affective Computing (AC) and Human-Robot Interaction (HRI), particularly in contexts related to wellbeing \cite{spitale2023vita}. Involving humans in experiments poses challenges and time constraints, making it even more difficult to collect a balanced dataset and in turn negatively impacting fairness.
This is also supported by literature that highlighted how the problem of bias is often associated with data imbalances \cite{chakraborty2021bias}. Imbalanced datasets can lead to bias against certain groups and that oversampling can improve model performance for underrepresented groups \cite{mehrabi2021survey}. \cite{pahl2022female} has annotated benchmark affective computing datasets to analyse the bias in detecting facial expressions. Their results show biases in age and ethnicity groups, for which models performed better with people under 34 years old. To overcome the issue of imbalanced datasets, many strategies have been proposed spanning from data augmentation, ensemble techniques, to evaluation metrics for imbalanced data.
Curating the dataset by ensuring balanced participation and representation across sensitive groups (e.g., gender, race) during data collection is crucial for fairness. When this is not applicable, data balancing techniques can be employed. Simple methods such as data augmentation via upsampling are already capable of significantly improving the lack of fairness present.
Therefore, we recommend to train human-centric ML models on balanced datasets by using simple balancing techniques such as data augmentation.

\textbf{R2: Make use of multimodal and higher level features.}
Our results show that fairness has been improved by using multi-modal and high-level features. 
Small AC and HRI datasets often include audio-visual recordings and sometimes physiological signals (e.g., EEG \cite{grissmann2017context}, heart rate variability, etc.). As such, they encompass different modalities that can be combined to provide more information for a machine learning algorithm to learn from.
Past works \cite{poria2017review, mathur2020introducing} have shown how multi-modal models performed better than uni-modal ML models. For example, \cite{song2020spectral} proposed two novel spectral representations, i.e., spectral heatmaps and spectral vectors, to represent video-level multi-scale temporal dynamics of expressive behaviour to detect depression from multi-modal data. Their results suggest that multi-modal models outperformed current state of the art in detecting depression. 
Conducting experiments with multiple modalities is crucial as they carry more information for the ML models to learn from. This should involve experiments with different fusion strategies (e.g., early or late fusion) to evaluate the best way to combine multi-modal information \cite{poria2017review}. 
Additionally, we suggest extracting high-level features from small datasets and selecting the most informative ones to train the models on, which can also improve fairness. 
Therefore, we recommend using high-level and multi-modal features when working with human-centric small datasets.

\textbf{R3: Employ a variety of ML models and evaluation and fairness metrics.}
Our results highlight the importance of exploring a variety of machine learning models and evaluating them using diverse evaluation and fairness metrics. This approach is a best practice to adopt when conducting ML experiments, and it is especially crucial when assessing fairness. Researchers can gain a more holistic understanding of the model's performance and potential biases when working with human-centric datasets, where fairness is a critical consideration.
Past works \cite{cheong_gender_fairness} emphasised the importance of using various measures for fairness to represent different aspects of bias. 
For instance, Equity-Accuracy (EA) assesses fairness by considering the model's predictive accuracy, whereas Disparate Impact (DI) evaluates fairness by focusing on the predicted outcomes.
In addition, a model that demonstrates strong fairness across diverse subgroups may exhibit lower accuracy compared to a model optimized solely for high performance. Therefore, it is crucial for each researcher to determine the appropriate fairness-accuracy trade off that best suits the specific task and context at hand \cite{Cheong_2023_WACV,churamani2023towards}.
Specific ML models may provide more favourable results across certain accuracy and fairness metrics. Therefore, studies should report on experimental results across a variety of metrics  (e.g., fairness-accuracy trade off). 
%
This will help ensure that the developed AI-based systems, like robotic coaches, are not only accurate, but also fair when deployed for delivering wellbeing practices.

\textbf{R4: Balance the trade off between AC/robotic system complexity and fairness.}
Our results show that the adaptive capability of robotic systems may have impacted the debiasing strategies in different ways. The complexity of adaptive capabilities can pose challenges in effectively enhancing fairness, while the adaptability of robotic systems might inherently reduce bias. 
Recently, the HRI field is rapidly employing  autonomous and adaptive robots that leverage AI components  (e.g., ChatGPT) and capabilities to create naturalistic and smooth robot-human interactions \cite{spitale2023vita}. These advancements led to improvement in the interaction outcomes and user perceptions towards the robot \cite{spitale2023vita}, but also to an increased complexity and reduced transparency of the robotic system, making it more difficult to control for bias.
This has been supported by the literature as well \cite{yan2020mitigating, wan2023biasasker}. 
For instance, Large Language Models pose challenges in debiasing due to their non-transparent and complex nature when embedded in robotic systems. They can exhibit emergent behaviours and unintended consequences that are difficult to predict and control for, causing further challenges when embedded into embodied artificial systems such as robots.

Despite the aforementioned challenges, adaptive models still need to be assessed in terms of bias and fairness before they can be embedded in embodied systems that will be deployed in real world settings with real world consequences \cite{ma2023fairness}.
When this is not feasible, real-time models should incorporate in-processing bias mitigation techniques. 
Therefore, we recommend researchers to carefully balance embodied / robotic system complexity and fairness to safeguard against biased robot-human interactions.

\textbf{R5: Define field and context specific ethical principles when designing / deploying AC systems for wellbeing.} 
Our results suggest that the use of robotic coaches may help improve fairness if, as in the case of the AFAR-RC22 \cite{spitale2023robotic} and AFAR-RC23 \cite{spitale2023vita, axelsson2024ohsorry}, the design of the robot-human interactions adhere to certain ethical principles\cite{axelsson2022robots}. For designing robotic mental wellbeing coaches, \cite{axelsson2022robots} distilled a set of design and ethical recommendations through an iterative process involving stakeholders, such as human wellbeing coaches and coachees.  When we collected the AFAR-RC22 and AFAR-RC23 datasets, we aimed to adhere to these guidelines when deploying robotic coaches to deliver positive psychology exercises in the workplace.
Other relevant works  \cite{londono2022fairness} also  highlight how the employment of ethical guideline may enhance fairness of AI-based systems \cite{chen2023ai, biondi2023ethical}. 
Recent EU guidelines also emphasise the use of a human-centric approach by involving stakeholders to define and discuss ethical considerations. In other research fields, there exists notable recent efforts in this direction. For example, \cite{biondi2023ethical} highlights the importance of ethical design in AI-based decision-making systems and \cite{chen2023ai} explores strategies to reduce bias and improve fairness in AI systems.
However, in the AC and HRI fields, no work has yet distilled a set of ethical guidelines to reduce bias in AI-based models and embodied systems like robots.

To ensure fair human-AI/robot interactions 
it is of utmost importance to adhere to ethical principles when taking the developed systems and robots out into the real world. Developing field and context specific guidelines for ethics and fairness will no doubt enable more purposeful and responsible innovation in AC and HRI research fields while expanding their research scope and significance.

\section*{Acknowledgments}
\small
\noindent\textbf{Funding:} J. Cheong is funded by the Alan Turing Institute Doctoral Studentship and the Leverhulme Trust. M. Spitale is supported by PNRR-PE-AI FAIR project funded by the NextGeneration EU program. H. Gunes is supported by the EPSRC/UKRI under grant ref. EP/R030782/1 (ARoEQ). \\
\textbf{Open Access:} For open access purposes, the authors have applied a Creative Commons Attribution (CC BY) licence to any Author Accepted Manuscript version arising.\\
\textbf{Data access:} Raw data related to this publication cannot be openly released due to anonymity and privacy issues. 

\small
\bibliographystyle{IEEEtran}

\bibliography{Micol,Jiaee}

\begin{IEEEbiographynophoto}{Jiaee Cheong} is a Turing doctoral student at the University of Cambridge.
Her research interests lie at the intersection of machine learning, affective computing, fairness, causality and HRI.
\end{IEEEbiographynophoto}
\begin{IEEEbiographynophoto}{Micol Spitale} is an Assistant Professor at the Department of Electronics, Information and Bioengineering at the Politecnico di Milano, as well as a Visiting Affiliated Researcher at the University of Cambridge.
In recent years, her research has been focused on the field of Social Robotics, Human-Robot Interaction, and Affective Computing, exploring ways to develop robots that are socio-emotionally adaptive and provide ‘coaching’ to promote wellbeing.
\end{IEEEbiographynophoto}
\begin{IEEEbiographynophoto}{Hatice Gunes} is a Full Professor of Affective Intelligence and Robotics (AFAR) in the Department of Computer Science and Technology, University of Cambridge, leading the \href{https://cambridge-afar.github.io/} {Cambridge AFAR Lab}. She is a former President of the Association for the Advancement of Affective Computing, a former Faculty Fellow of the Alan Turing Institute and is currently a Fellow of the EPSRC and Staff Fellow of Trinity Hall.
\end{IEEEbiographynophoto}

\end{document}